\documentclass[10pt,twocolumn,letterpaper]{article}

\usepackage{iccv}
\usepackage{times}
\usepackage{url}
\usepackage{epsfig}
\usepackage{graphicx}
\usepackage{amsmath}
\usepackage{amssymb}
\usepackage{amsmath,epsfig}
\usepackage{pdfpages}
\usepackage{amsfonts}
\usepackage{pifont}
\usepackage{algorithm}
\usepackage{graphicx} 
\usepackage[noend]{algpseudocode}
\usepackage{multirow}
\usepackage{color}
\usepackage{tabularx}
\usepackage{makecell}
\usepackage{cite}
\usepackage{amssymb}
\usepackage{subcaption}
\usepackage{authblk}
\definecolor{red}{rgb}{1,0,0}
\definecolor{blue}{rgb}{0,0,1}

\newcommand{\cmark}{\ding{51}}%
\newcommand{\xmark}{\ding{55}}%
\usepackage[pagebackref=true,breaklinks=true,bookmarks=false, colorlinks]{hyperref}

\iccvfinalcopy 

\def\iccvPaperID{6740} 

\ificcvfinal\fi

\begin{document}

\title{Cascaded Parallel Filtering for Memory-Efficient Image-Based Localization}
\author{Wentao Cheng$^{1,2}$, Weisi Lin$^1$, Kan Chen$^2$ and Xinfeng Zhang$^3$\\ $^1$Nanyang Technological University, Singapore $^2$Fraunhofer Singapore, Singapore\\ 
       $^3$University of Chinese Academy of Sciences, China\\\vspace{+0.2em} \tt\small wcheng005@e.ntu.edu.sg, wslin@ntu.edu.sg, chen.kan@fraunhofer.sg, zhangxinf07@gmail.com}
\maketitle
\ificcvfinal\thispagestyle{empty}\fi

\begin{abstract}
Image-based localization (IBL) aims to estimate the 6DOF camera pose for a given query image. The camera pose can be computed from 2D-3D matches between a query image and Structure-from-Motion (SfM) models. Despite recent advances in IBL, it remains difficult to simultaneously resolve the memory consumption and match ambiguity problems of large SfM models. In this work, we propose a cascaded parallel filtering method that leverages the feature, visibility and geometry information to filter wrong matches under binary feature representation. The core idea is that we divide the challenging filtering task into two parallel tasks before deriving an auxiliary camera pose for final filtering. One task focuses on preserving potentially correct matches, while another focuses on obtaining high quality matches to facilitate subsequent more powerful filtering. Moreover, our proposed method improves the localization accuracy by introducing a quality-aware spatial reconfiguration method and a principal focal length enhanced pose estimation method. Experimental results on real-world datasets demonstrate that our method achieves very competitive localization performances in a memory-efficient manner.
\end{abstract}
\begin{table*}[!t]
	\small
	\begin{center}
		\begin{tabular}{lcccccccc}
			\Xhline{1pt}
			\multirow{2}{*}{Method} & \multirow{2}{*}{Feature Type} & \multirow{2}{*}{Compactness} & \multicolumn{3}{c}{Match Filtering} & \multirow{2}{*}{Prior-free} & \multirow{2}{*}{SR} \\ \cline{4-6} 
			&  &  & Feature-wise & Visibility-wise & Geometry-wise & & \\ \hline
			AS \cite{sattler2012improving} & SIFT & \xmark & Strict &  \cmark  & \xmark & \cmark & \xmark  \\ \hline
			WPE \cite{li2012worldwide} & SIFT & \xmark & Relaxed &  \cmark  & \xmark & \cmark & \xmark \\ \hline
			CSL \cite{svarm2017city} & SIFT & \xmark & Relaxed & \xmark & \cmark & \xmark$^{*}$ & \xmark  \\ \hline
			CPV \cite{zeisl2015camera} & SIFT & \xmark & Relaxed & \xmark & \cmark & \xmark$^{*}$ & \xmark \\ \hline
			Hyperpoints \cite{sattler2015hyperpoints} & SIFT & \cmark & Relaxed &  \cmark  & \cmark & \cmark & In RPE  \\ \hline
			EGM \cite{liu2017efficient} & SIFT+Binary & \xmark & Relaxed &  \cmark  & \xmark & \cmark & \xmark\\ \hline
			TC \cite{camposeco2017toroidal} & SIFT & \xmark & Relaxed & \xmark & \cmark & \cmark & \xmark \\ \hline
			SMC \cite{toft2018semantic} & SIFT & \xmark & Relaxed & \cmark & \cmark & \xmark$^{*}$ & \xmark \\ \hline
			Our method & Binary & \cmark & Relaxed &  \cmark  & \cmark & \cmark  & Before RPE  \\ \hline
		\end{tabular}
	\end{center}
	\vspace{-1.5em}
	\caption{Comparison between our method and other structure-based IBL methods. \xmark$^{*}$ means that the vertical direction of camera is known in advance, SR represents Spatial Reconfiguration and RPE represents RANSAC-based Pose Estimation.} \label{table::summary}
	\vspace{-1.4em}
\end{table*}
\section{Introduction}
Image-based localization (IBL), \ie computing the 6DOF camera pose for a query image, is a fundamental problem in many computer vision tasks. For example, IBL plays a key role in incremental Structure-from-Motion (SfM) reconstruction \cite{snavely2006photo,goesele2007multi}, visual place recognition \cite{sattler2015hyperpoints}, and visual navigation for autonomous vehicles \cite{sattler2018benchmarking}. IBL has witnessed tremendous advancement by means of deep learning \cite{kendall2015posenet,kendall2017geometric} and image retrieval techniques \cite{arandjelovic2013all,arandjelovic2016netvlad,sattler2017large}. However, structure-based IBL \cite{li2012worldwide,sattler2012improving,zeisl2015camera,svarm2017city,liu2017efficient,camposeco2017toroidal,toft2018semantic} by directly establishing 2D-3D matches between a query image and SfM models is still the most prevailing strategy. Recent state-of-the-art methods handle the match ambiguity under high-dimensional feature representation with semantic consistency \cite{toft2018semantic}. However, it remains challenging and crucial to solve this problem under compact feature representation.

A large SfM model requires prohibitive memory consumption to store tens of millions of descriptors. Meanwhile, match filtering becomes difficult, as it may contain many nearly identical descriptors. Particularly, the feature (\eg visual similarity), visibility (\eg point-image relationship), and geometry (\eg camera pose) information in IBL leads to three interesting questions: Is it possible to improve the discriminative power of each information? How to unify them so that each can play its proper role, \ie use its discriminative power to a tee? When is the appropriate phase to engage one specific information in an IBL pipeline? The accuracy is also a key issue for IBL especially in autonomous driving applications. The camera pose can be estimated by using a minimal pose solver \cite{bujnak2008general} in RANSAC \cite{fischler1981random}. To achieve high accuracy, degenerate pose hypotheses should be prevented from being sampled or selected.

In this paper, we propose a cascaded parallel filtering method with respect to a binary feature representation via Hamming Embedding \cite{jegou2008hamming}. Using this binary feature representation, we can largely reduce the memory consumption. Meanwhile, it will introduce more ambiguities than high-dimensional feature representation, making match filtering notoriously harder. To break this dilemma, our proposed method filters wrong matches in a cascaded manner by sequentially leveraging the intrinsic feature, visibility, and geometry information. When engaging one type of information, we use a relaxed criterion to reject matches and retain a match pool that focuses on preserving correct matches. In parallel, we use a strict criterion to obtain high confident matches, which facilitate subsequent filtering steps. In feature-wise filtering, we reformulate a traditional match scoring function \cite{jegou2009burstiness} with a bilateral Hamming ratio test to better evaluate the distinctiveness of matches. In visibility-wise filtering, we explore the point-image relationship to filter wrong matches by retrieving relevant database images. Moreover, we propose a two-step match selection method by exploring the point-point relationship, which allows us to obtain substantial 2D-3D matches for computing an auxiliary camera pose. In geometry-wise filtering, we apply this auxiliary camera pose on the retained match pool to reject wrong matches by means of re-projection error.

Our method also aims to improve the localization accuracy based on two key observations. The first observation is that, correct matches that appear in sparse regions, are essential to establish a non-degenerate camera pose hypothesis. Due to the scarcity of such matches, they are usually neglected in camera pose estimation. Consequently, we propose a quality-aware spatial reconfiguration method to increase the possibility of sampling such matches in RANSAC-based pose estimation. The second observation is that, several top ranked camera pose hypotheses that have similar and realistic focal length values, are more robust than the camera pose hypothesis with the largest number of inliers. Based on this, we shift the focus to find a principal focal length value so that we can obtain a more accurate camera pose accordingly. The evaluation on benchmark datasets demonstrates that our method gets promising localization results with significantly lower memory consumption comparing with state-of-the-art methods. The source code of our method is available at \url{https://github.com/wentaocheng-cv/cpf_localization} 

\noindent \textbf{Related work.} In recent years, numerous structure-based IBL approaches \cite{irschara2009structure,li2010location,sattler2011fast,li2012worldwide,choudhary2012visibility,sattler2012improving,sattler2012image,cao2014minimal,sattler2015hyperpoints,sattler2017efficient,zeisl2015camera,svarm2017city,camposeco2017toroidal,cheng2017data,liu2017efficient,tran2018device} have been proposed. Table \ref{table::summary} shows an overview of state-of-the-art structure-based IBL approaches. Feature-wise filtering that mainly relies on the widely used SIFT ratio test \cite{lowe2004distinctive} is a fundamental strategy in IBL. Efficient 2D-3D feature matching methods \cite{choudhary2012visibility,sattler2012improving} require a strict feature-wise filtering criterion to generate highly confident seed matches. The matches that are co-visible frequently with seed matches are prioritized to accelerate the matching process. Recent works \cite{li2012worldwide,zeisl2015camera, svarm2017city, camposeco2017toroidal,toft2018semantic,Sarlin2018Leveraging} commonly relax the feature-wise filtering criterion to preserve more correct matches and shift the filtering task to visibility or geometry tools. Li \etal introduce a RANSAC sampling strategy by prioritizing samples with frequent co-visibility \cite{li2012worldwide}. Liu \etal propose a ranking algorithm by globally exploiting the visibility information on a Markov network \cite{liu2017efficient}. Top ranked matches are then filtered through traditional SIFT ratio test. Camposeco \etal propose a geometric outlier filtering approach, in which a novel 2-point solver is able to compute an approximate camera position \cite{camposeco2017toroidal}. Assuming that the gravity direction and an approximate estimation of camera height are known, both Zeisl \etal and Svarm \etal present geometric outlier filtering approaches to handle extremely large outlier ratios \cite{camposeco2017toroidal,svarm2017city}. Toft \etal derive an outlier filtering method by combining the known gravity direction prior and semantic information \cite{toft2018semantic}.

In order to reduce the memory consumption of large SfM models, point cloud simplification approaches \cite{li2010location,cao2014minimal,cheng2017data,Lynen2015Get} select a subset of representative 3D points by formulating a set cover problem. However, the reduction of points usually decreases the localization effectiveness and accuracy. Learning-based approaches implicitly compress the SfM model by training a CNN model to regress the camera pose \cite{kendall2015posenet,kendall2017geometric,walch2017image} or scene coordinates \cite{brachmann2018learning}. Yet, when facing large SfM models, these approaches either have low accuracy \cite{kendall2015posenet,kendall2017geometric} or encounter a complete training failure \cite{brachmann2018learning}. Sattler \etal quantize the model descriptors into a 16M fine visual vocabulary to reduce memory consumption \cite{sattler2015hyperpoints}. To handle the ill-conditioned spatial distribution, they improve the effective inlier count algorithm \cite{irschara2009structure} and apply it in the RANSAC verification stage. In contrast, our proposed quality-aware spatial reconfiguration method is employed before RANSAC-based pose estimation, which allows us to obtain more non-degenerate pose hypotheses with the same number of RANSAC iterations. 
\begin{figure*}[!t]
	\centering
	\includegraphics[width=\textwidth]{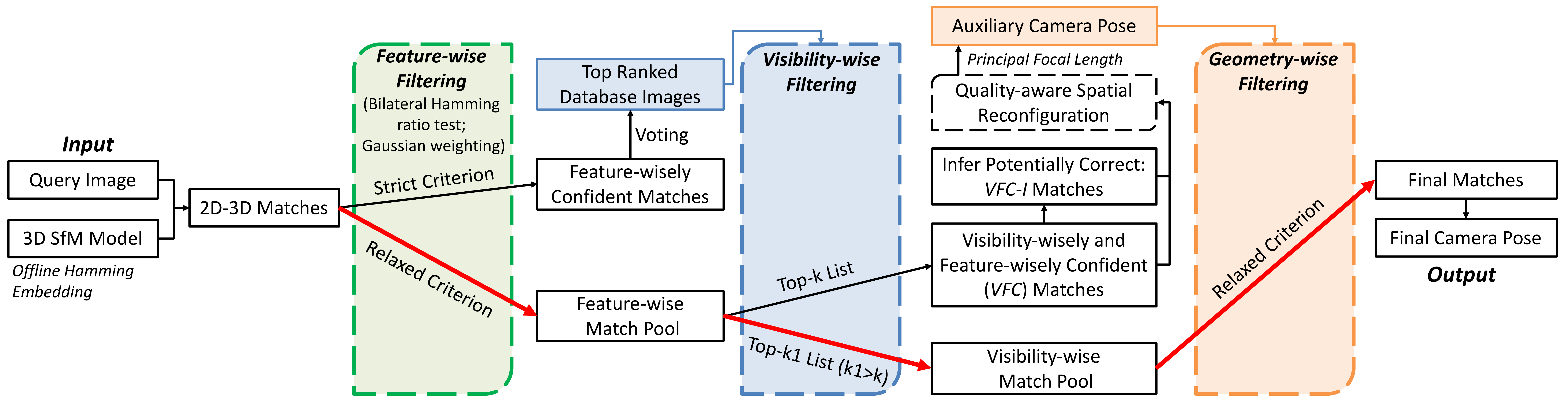}
	\caption{Overview of the localization pipeline using our cascaded parallel filtering method. A cascade of feature-, visibility-  and geometry-wise filtering steps is illustrated from left to right.  Two parallel tasks are applied in both the feature-wise (green) and visibility-wise (blue) filtering steps. One task with strict criterion aims for facilitating subsequent steps. Another task with relaxed criterion (in red arrow) aims for preserving correct matches. } \label{fig::overview}
	\vspace{-1.3em}
\end{figure*}
\section{Proposed Method}
Fig. \ref{fig::overview} shows the structured-based IBL pipeline using our method. In this section, we describe each step in detail.
\subsection{Feature-wise Match Filtering} \label{sec::evaluation_match} 
First, we introduce the feature-wise match filtering step. The goal of this step is twofold: 1) to retain a feature-wise match pool by rejecting obviously wrong matches, 2) to obtain a set of feature-wisely confident matches to facilitate subsequent filtering steps. \\ \vspace{-0.8em}

\noindent \textbf{Data pre-processing. }Let $\mathcal{P}$ be the 3D points in a SfM model. Each 3D point is associated with a set of SIFT descriptors. A general or specific visual vocabulary should firstly be trained using clustering techniques. In the offline stage, the descriptors of a 3D point are assigned to their closest visual words through nearest neighbor search. For efficiency, we follow \cite{sattler2012improving} by representing the SIFT descriptors of a 3D point as an integer mean descriptor per visual word. Subsequently, each integer mean descriptor is converted into a compact binary signature containing $B$ bits using Hamming Embedding \cite{jegou2008hamming}. Given a query image, a set of SIFT descriptors are extracted, denoted as $\mathcal{Q}$. For each descriptor $q \in \mathcal{Q}$, we first assign it to its closest visual word. Using Hamming Embedding, we also obtain the binary signature for descriptor $q$, denoted as $s_{q}$. For each 3D point $p \in \mathcal{P}$, if one of its associated integer mean descriptors is quantized into the same visual word with query descriptor $q$, a 2D-3D match can be established as $m =\left \{q \leftrightarrow p \right \}$. The Hamming distance of $m$ can be measured as $h(s_{q},s_{p})$. \\ \vspace{-0.8em}

\noindent \textbf{Bilateral Hamming ratio test. }To evaluate the distinctiveness of the resultant 2D-3D matches, previous works mainly focus on the SfM model side by using a fixed Hamming distance threshold \cite{sattler2012image}, Gaussian weighting \cite{jegou2012aggregating}, or density estimation \cite{arandjelovic2014dislocation}. Few attentions have been paid on filtering on the query image side, where the corresponding feature space is easier to distinguish correct matches due to its sparsity. Inspired from the variable radius search in \cite{zeisl2015camera}, we propose a bilateral Hamming ratio test that operates on both the query image and the SfM model. 

In order to prevent correct matches from being rejected in this step, we apply a coarse filtering scheme by using a large Hamming distance threshold $\tau$. Therefore, for a match $m = \left \{ q \leftrightarrow p \right \}$, the set of 3D points that can form a match with query descriptor $q$ can be defined as $\mathcal{P}(q) = \left \{ p\in \mathcal{P} | h(s_{q},s_{p}) \leq \tau \right \}$. Similarly, the set of query descriptors that can form a match with 3D point $p$ can be represented as $\mathcal{Q}(p) = \left \{ q\in \mathcal{Q} | h(s_{q},s_{p}) \leq \tau \right \}$. Our core idea is that a match should be distinctive if its corresponding Hamming distance is significantly lower than the average Hamming distance in $\mathcal{P}(q)$ and $\mathcal{Q}(p)$. To evaluate a match within the feature space of a query image, we apply an image side Hamming ratio test as follows:
\begin{equation} 
t(m) = \frac{\sum_{j\in \mathcal{Q}(p)} h(s_{j},s_{p}) }{h(s_{q},s_{p}){\left|\mathcal{Q}(p)  \right |}^{2}},
\end{equation}
where one $\left|\mathcal{Q}(p)  \right |$ in ${\left|\mathcal{Q}(p)  \right |}^{2}$ is used to compute the average Hamming distance, and another is to penalize a match whose corresponding 3D point establish multiple matches. It is safe to reject a match when it is obviously ambiguous in the feature space of a query image. Therefore, we reject matches if their corresponding image side ratio test scores are smaller than a threshold $\varphi$. We observe that setting $\varphi$ to $0.3$ works well in practice.

Similarly, to evaluate the distinctiveness of a match within the feature space of a SfM model, we apply the model side Hamming ratio test as follows:
\begin{equation}
{t}'(m) = \frac{ \sum_{j\in \mathcal{P}(q)} h(s_{q},s_{j})}{h(s_{q},s_{p}) \left|\mathcal{P}(q)  \right |}.
\end{equation}
Since the term $\left|\mathcal{P}(q)  \right |$ may vary dramatically with using different size of visual vocabularies, here we don't use it to penalize a match whose corresponding query descriptor can establish multiple matches with different 3D points. In addition, a large SfM model usually contains orders of magnitude more descriptors than an image. This makes the model side Hamming ratio test prone to reject correct matches by directly setting a hard threshold. Therefore, we only apply ${t}'(m)$ as a soft scoring function to evaluate a match. The final bilateral Hamming ratio test can be defined as follows:

\begin{equation} \label{eq::ratio_test}
T(m) = \left\{ \begin{array}{lr}
{t}'(m), & {t}(m) \geq \varphi \\
0, & \text{otherwise}. \end{array} \right. 
\end{equation}\\ \vspace{-1em}

\noindent \textbf{Aggregating Gaussian weighting function. }In order to strengthen the feature distinctiveness, we propose an adapted version of Gaussian weighting function \cite{jegou2009burstiness} as follows:
\begin{equation} \label{eq::final_score}
w(h) = \left\{ \begin{array}{lr}
{(\frac{\sigma}{h})}^{2}e^{-{(\frac{h}{\sigma})}^{2}}, & 0.5\sigma < h \leq \tau \\
4e^{-0.25}, & 0 < h \leq 0.5\sigma \\
0, & \text{otherwise}, \end{array} \right.
\end{equation}
where $h$ is the Hamming distance of a match, and $\sigma$ is usually set to one quarter of the binary feature dimension \cite{arandjelovic2014dislocation}. By aggregating the Gaussian weighting function, the score for a match $m$ therefore can be computed as follows:
\begin{equation}\label{eq::evaluation}
E(m) = T(m)w(h(m)).
\end{equation}
Overall, we can retain a feature-wise match pool $\mathcal{M} = \left \{ m | E(m) > 0 \right \}$, which focuses on preserving correct matches. We also obtain a set of Feature-wisely Confident (\emph{FC}) matches $\mathcal{M}_{FC} = \left \{ m | E(m) \geq \alpha \right \}, \alpha > 0$.

\subsection{Visibility-wise Match Filtering} \label{section:visibility}
Given the match sets $\mathcal{M}$ and $\mathcal{M}_{FC}$, we describe how to leverage the visibility information in a SfM model to further filter wrong matches. In particular, we aim to achieve two purposes at this stage: 1) to reject wrong matches in $\mathcal{M}$ to retain a visibility-wise match pool that well preserves correct matches, 2) to select a set of high quality matches that are substantial to derive an auxiliary camera pose for later geometry-wise filtering. The visibility information encoded in a SfM model can be represented as a bipartite visibility graph $\mathcal{G} = \left \{ \mathcal{P},\mathcal{D},\mathcal{E} \right \}$. Each node $p \in \mathcal{P}$ represents a 3D point, and each node $d \in \mathcal{D}$ represents a database image. An edge $(p,d) \in \mathcal{E}$ exists if point $p$ is observed in database image $d$. Intuitively, correct matches usually cluster in the database images that are relevant to a given query image. Thus, the problem of match filtering can be transferred as a problem of finding relevant database images. \\ \vspace{-0.8em}

\noindent \textbf{Voting with \emph{FC} matches.} Using the visibility graph $\mathcal{G}$, a 2D-3D match $m =\left \{ q \leftrightarrow p \right \}$ can cast a vote to each database image that observes point $p$. In order to prevent ambiguous matches from interfering the voting procedure, we only use \emph{FC} matches to vote database images. Inspired from \cite{sattler2015hyperpoints}, we also enforce a locally unique voting scheme. Let $\mathcal{M}^{d}_{FC} = \left \{ m=\left \{ q\leftrightarrow p \right \}| m\in \mathcal{M}_{FC}, (p,d) \in \mathcal{E} \right \}$ be the \emph{FC} matches that vote for database image $d$. We enforce that a match for database image $d$ can be added to $\mathcal{M}^{d}_{FC}$ only if its corresponding query descriptor has not appeared in $\mathcal{M}^{d}_{FC}$ before. In addition, we only consider database images that receive at least three votes to ensure high relevancy to the query image. After accumulating the match scores for a database image, we adopt a term frequency weight in order to penalize database images that observe a large number of 3D points. Let $\mathcal{P}^{d} = \left \{ p | (p,d) \in \mathcal{E} \right \}$ be the set of 3D points that are observed by the database image $d$, the voting score can be defined as follows: 
\begin{equation}\label{eq:voting}
\mathcal{S}(d) = \frac{\sum_{m \in \mathcal{M}^{d}_{FC}}E(m)}{\sqrt{\left | \mathcal{P}^{d} \right |}}.
\end{equation}
A larger voting score inherently indicates that the corresponding database image is more relevant to a given query image, hence more likely to find correct matches. We first retrieve top-$k$ ranked database images $d(k)$ with the largest voting scores. For a match $m \in \mathcal{M}$, we select it into the set $\mathcal{M}^{d(k)}$ if its corresponding 3D point is observed in at least one of the images in $d(k)$. Note that only visibility information is considered and we preserve both \emph{FC} and \emph{non-FC} matches in $\mathcal{M}^{d(k)}$. Similarly, we apply a relaxed criterion by using a larger $k_{1}$ to select another set of matches $\mathcal{M}^{d(k_{1})}$, which may contain more correct matches but also are more noisy than $\mathcal{M}^{d(k)}$. $\mathcal{M}^{d(k_{1})}$ will serve a visibility-wise match pool and later be filtered in Section \ref{Section:geometry}. \\  \vspace{-0.8em}

\noindent \textbf{Two-step match selection.} Naturally, we can define the matches in $\mathcal{M}^{d(k)}$ as \emph{Visibility-wisely Confident (VC)} matches. Due to the existence of feature-wisely ambiguous matches, \emph{VC} matches may contain a large portion of outliers, making them difficult to be directly applied in camera pose estimation. We propose a two-step match selection method to filter \emph{VC} matches. In the first step, we select the \emph{FC} from \emph{VC} matches as \emph{Visibility-wisely and Feature-wisely Confident (VFC)} matches that can be defined as follows:
\begin{equation}\label{eq:vfc}
\mathcal{M}^{d(k)}_{VFC} = \left \{ m |m \in \mathcal{M}^{d(k)}  \wedge E(m) \geq \alpha  \right \}.
\end{equation}
The \emph{VFC} matches exhibit high confidence to be correct since they not only are observed in top ranked database images, but also are highly distinctive in feature space. The major difficulty is how to distinguish correct matches from the rest \emph{Visibility-wisely but Not Feature-wisely Confident (VNFC)} matches that can be defined as $\mathcal{M}^{d(k)}_{VNFC} = \mathcal{M}^{d(k)} \setminus \mathcal{M}^{d(k)}_{VFC}$.

During the image voting procedure, we leverage the point-image relationship in the bipartite visibility graph $\mathcal{G}$. Now we use the point-point relationship in $\mathcal{G}$ to help us filter the \emph{VNFC} matches. Intuitively, if a 3D point of one \emph{VNFC} match exhibits a strong co-visibility relationship with 3D points of \emph{VFC} matches in top ranked database images, it should be regarded as a potentially correct match. To this end, we engage the second step match selection to infer potentially correct matches from \emph{VNFC} matches. For each database image $d \in d(k)$, we first count the number of \emph{VFC} matches and \emph{VNFC} matches, which we call ${\omega}_{VFC}^{d}$ and $\omega_{VNFC}^{d}$ respectively. If \emph{VFC} matches occupy a larger portion compared with \emph{VNFC} matches in one database image, each \emph{VNFC} match should receive stronger promotion from \emph{VFC} matches respectively. Therefore, for an \emph{VNFC} match, we compute its updated match score as follows:
\begin{equation}\label{eq:update}
E^{'}(m) = E(m) + \sum_{d \in d(k)}\frac{\alpha}{2}\ln(1 + \frac{ {\omega}_{VFC}^{d}}{ {\omega}_{VNFC}^{d}}).
\end{equation}
The larger the updated match score, the more likely that corresponding \emph{VNFC} match is correct. Using the previous match score threshold $\alpha$, we can select a set of potentially correct matches from \emph{VNFC} matches. Since these potentially correct matches are mainly inferred by exploring the visibility information with \emph{VFC} matches, we call them \emph{VFC-I} matches and they can be defined as follows:
\begin{equation}
\mathcal{M}^{d(k)}_{VFC-I} = \left \{ m | m \in \mathcal{M}^{d(k)}_{VNFC} \wedge E^{'}(m) \geq \alpha\right \}. 
\end{equation}
Therefore, the matches that we select from $\mathcal{M}^{d(k)}$ are the union of \emph{VFC} and \emph{VFC-I} matches. Algorithm \ref{algo:visibility} illustrates the process of visibility-wise match filtering.

\begin{algorithm} [!t]%
	\caption{Visibility-wise Match Filtering}
	\label{algo:visibility}
	\begin{algorithmic}[1]
		\Require Matches $\mathcal{M}$ with feature-wise match scores $E(m)$, match score threshold $\alpha$
		\Require  $\mathcal{M}_{VFC}^{d(k)} \gets \emptyset$, $\mathcal{M}_{VFC{\text -}I}^{d(k)} \gets \emptyset$,$\mathcal{M}^{d(k_1)} \gets \emptyset$
		\State {/* explore \textbf{point-image } visibility */} 
		\State {Apply image voting with \emph{FC} matches using Eq. \ref{eq:voting}} 
		\State {Retrieve top $k$ and $k_1$ database images $d(k)$ and $d(k_1)$}
		\State {Select all matches in $d(k_1)$ as $\mathcal{M}^{d(k_1)}$ for visibility-wise match pool}
		\State {Separate \emph{VFC} matches $\mathcal{M}_{VFC}^{d(k)}$ and \emph{VNFC} matches $\mathcal{M}_{VNFC}^{d(k)}$ using Eq. \ref{eq:vfc}}
		\State {/* explore \textbf{point-point} visibility */} 
		\ForAll {$d \in d(k)$}
		\State {Compute the number of \emph{VFC} matches ${\omega}_{VFC}^{d}$} 
		\State {Compute the number of \emph{VNFC} matches ${\omega}_{VNFC}^{d}$} 
		\ForAll {$m \in \mathcal{M}_{VNFC}^{d} $}
\State {Compute the updated match score $E^{'}(m)$ using Eq. \ref{eq:update}}  
		\EndFor
		\EndFor
		\ForAll {$m \in \mathcal{M}_{VNFC}^{d(k)}$} 
		\If {$E^{'}(m) \geq \alpha$}
		\State $\mathcal{M}_{VFC{\text -}I}^{d(k)} \gets \mathcal{M}_{VFC{\text -}I}^{d(k)}\cup\left \{ m \right \}$
		\EndIf
		\EndFor
		\State \Return $\mathcal{M}_{VFC}^{d(k)} \cup\mathcal{M}_{VFC{\text -}I}^{d(k)}$ and $\mathcal{M}^{d(k_1)}$
	\end{algorithmic}
\end{algorithm}
\subsection{Geometry-wise Match Filtering}\label{Section:geometry}
In this section, we describe how to use the obtained \emph{VFC} and \emph{VFC-I} matches to compute an auxiliary camera pose, which facilitates geometry-wise match filtering for the visibility-wise match pool $\mathcal{M}^{d(k_1)}$. \\  \vspace{-0.8em}

\begin{figure}[!t]
	\centering
\includegraphics[width=0.45\textwidth]{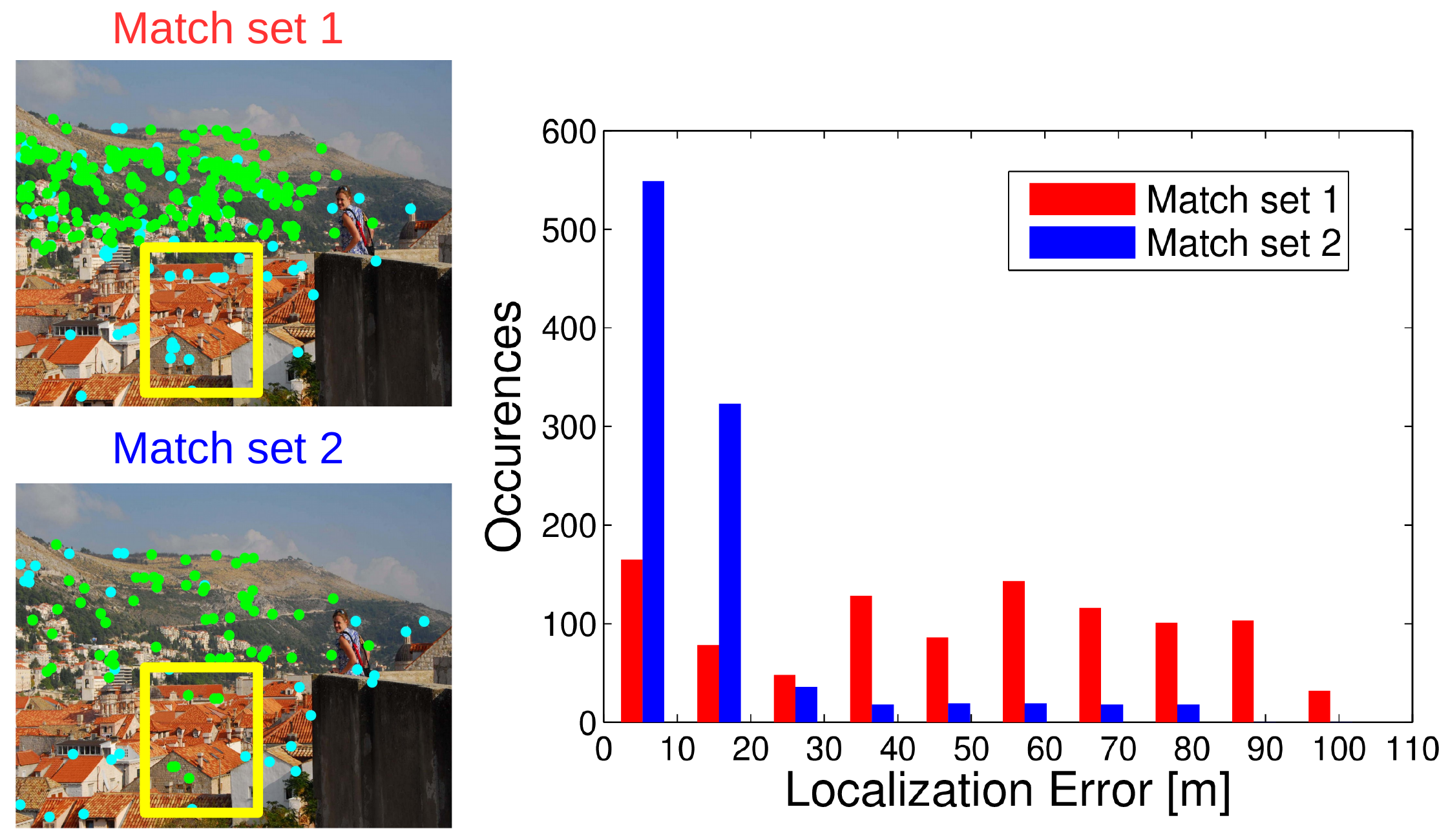}
	\caption{The influence of a uniform spatial distribution for matches. \textbf{Left top:} Original match set with 242 inliers shown in green and 64 outliers shown in cyan (inlier ratio is 0.79), matches are clustered in mountain area; \textbf{Left bottom:} a selection from original match set by applying spatial reconfiguration, this selection has 63 inliers and 31 outliers (inlier ratio is 0.67), matches are more uniformly distributed over the image; \textbf{Right:} Localization error statistics with these two match sets by running 1000 camera pose estimation trials. \textbf{Yellow box:} the inside correct but sparse matches are emphasized in match set 2.}\label{fig::example}
	\vspace{-1.8em}
\end{figure}

\noindent \textbf{Quality-aware spatial reconfiguration.} A common way to estimate the camera pose is to use pose solvers inside RANSAC loops. The quality of input 2D-3D matches, \ie the inlier ratio, is an essential factor for robust and efficient camera pose estimation. It is also important to ensure that the input matches have a uniform spatial distribution, especially when the majority of input matches cluster in a highly textured region as shown in Fig. \ref{fig::example}. Correct matches, rare but critical, in poorly textured regions are unlikely to be sampled in the RANSAC hypothesis stage. This will significantly reduce the localization accuracy due to the difficulty of obtaining a non-degenerate pose hypothesis. 

Our goal is to obtain a set of matches that simultaneously have a large inlier ratio and a uniform spatial distribution by selecting from \emph{VFC} and \emph{VFC-I} matches. To this end, we first divide the query image into $4$ by $4$ equally-sized bins, denoted as $\mathcal{B}$. The \emph{VFC} and \emph{VFC-I} matches are then quantized into $\mathcal{B}$ according to the image coordinates of their associated 2D query descriptors. To make the spatial distribution of selected matches more uniform, we apply a spatial reconfiguration method to penalize dense bins with more quantized matches and emphasize sparse bins with fewer quantized matches. Let $N_{b}$ be the number of matches that are quantized into bin $b \in \mathcal{B}$. Let $R_{b}$ be the proportion of matches that can be selected from bin $b$, the spatial reconfiguration can be realized by computing $R_{b}$ as follows:
\begin{equation}\label{eq::reconfiguration}
R_{b} = \frac{{\sqrt{N_{b}}}}{\sum_{i \in \mathcal{B}}{\sqrt{N_{i}}}}.
\end{equation}
To achieve an efficient camera pose estimation, we limit that overall at most $N$ matches can be selected. Accordingly, for each bin $b$, the match selection quota is $R_{b}N$. 

We first select the \emph{VFC} matches with larger match scores according to each bin's selection quota. After that, if there exist bins that still do not reach the selection quotas, we then select the \emph{VFC-I} matches from these bins. Note that the \emph{VFC-I} matches exhibit inferior quality than the \emph{VFC} matches because of their confidence in only visibility. To ensure high quality of selected matches, the \emph{VFC} matches should be dominant. Suppose the number of selected \emph{VFC} matches is $N_{VFC}$, we restrict that at most $\beta N_{VFC}$ \emph{VFC-I} matches can be selected. In this work, we set $\beta$ to $0.33$. \\  \vspace{-0.8em}

\noindent \textbf{Auxiliary camera pose with principal focal length.} We then use the selected matches after quality-aware spatial reconfiguration to compute an auxiliary camera pose. Assuming a general scenario when the focal length of a given query image is unknown, we can adopt a 4-point pose solver (P4P) \cite{bujnak2008general} to estimate the extrinsic calibration and the focal length. In RANSAC-based camera pose estimation, the estimated camera pose usually is the pose hypothesis that is supported by the largest number of inliers. However, we notice that this strategy becomes unreliable when few correct matches exist. In such case, a co-planar degenerated sample may result in that the estimated camera will lie far away from the scene with an unrealistic focal length. 
To tackle this unreliability problem, we propose a statistical verification scheme to find a reliable camera pose. Let $\varepsilon$ be the largest number of inliers of a pose hypothesis after running a certain number of RANSAC+P4P loops. We store the top-10 pose hypotheses, whose corresponding inliers are more than $0.7\varepsilon$. For a successful localization, we notice that most of the top hypotheses have numerically close focal length values. These focal length values, instead of the one with largest number of inliers, provide us a more stable and reliable camera pose estimation. Inspired from RANSAC variants \cite{optimal_ransac} that vote for optimal parameter values, we propose to select the pose hypothesis whose focal length is the median value among the top pose hypotheses. We define the selected pose hypothesis as an auxiliary camera pose, and its corresponding focal length as principal focal length $f$. \\ \vspace{-0.8em}

\noindent \textbf{Filtering with auxiliary camera pose.} The computed auxiliary camera pose exhibits sufficient accuracy. Using it to recover potentially correct matches back can further improve the localization accuracy. We apply the auxiliary camera pose on the visibility-wise match pool $\mathcal{M}^{d(k_1)}$ to realize the geometry-wise filtering. We define a relaxed re-projection error threshold $\theta$ in case rejecting potentially correct matches. As such, a match can be selected as a potentially correct match if the re-projection error with respect to the auxiliary camera pose is below $\theta$. In this work, we choose a threshold of 10 pixels.\\ \vspace{-0.8em}

\noindent \textbf{Final camera pose estimation.} The matches selected by the auxiliary camera pose exhibit both high quality and high quantity. In addition, we have also obtained a reliable focal length value $f$. Based on these, we can directly apply a 3-point pose solver (P3P) \cite{kneip2011novel}, which is much more efficient than 4-point pose solvers, to compute the final camera pose. 

\section{Experiments}\label{experiment}
\subsection{Datasets and Evaluation Metrics}
We evaluate our proposed method on four benchmark datasets as summarized in Table \ref{table:dataset}. For the Dubrovnik dataset, we adopt the same evaluation metric used in related works \cite{li2010location,sattler2011fast,sattler2012improving,zeisl2015camera,svarm2017city,camposeco2017toroidal,liu2017efficient}. A query image is considered as successfully registered or localized if the best camera pose after RANSAC has at least 12 inliers. The localization accuracy on the Dubrovnik dataset can be measured as the distance between estimated camera center position and the ground truth camera center position of query image. The RobotCar Seasons \cite{sattler2018benchmarking} dataset was reconstructed from images that were captured with cameras mounted on an autonomous vehicle. This dataset covers a wide range of condition changes, \eg weather, seasons, day-night, which make image-based localization on this dataset challenging. The ground truth camera poses of query images were obtained by aligning all 49 SfM sub-models to LIDAR point clouds. The query images of the Aachen Day-Night dataset consist of 824 images in day condition and 98 images in night condition. For the RobotCar Seasons and Aachen Day-Night datasets, we follow the evaluation metric in \cite{sattler2018benchmarking} and report the percentage of query images localized within $U$m and $V^{\circ}$ from ground truth camera poses. To evaluate under different levels of localization accuracy, we use the three accuracy intervals defined in \cite{sattler2018benchmarking} as follows: High-precision ($0.25m,2^{\circ}$), Medium-precision ($0.5m,5^{\circ}$) and Coarse-precision ($5m,10^{\circ}$). For the large-scale SF-0 dataset \cite{li2012worldwide}, we use the evaluation package provided by \cite{sattler2017large} which contains reference camera poses for 442 query images.
\begin{table}[!t]
	\centering
	\small
	\caption{Summarization of the used datasets.} \label{table:dataset}
	\vspace{-0.5em}
	\begin{tabular}{lccc}
		\Xhline{1pt}
		Dataset & \begin{tabular}[c]{@{}c@{}}Database \\ Images\end{tabular} & \begin{tabular}[c]{@{}c@{}}3D \\ Points\end{tabular} & \begin{tabular}[c]{@{}c@{}}Query \\ Images\end{tabular} \\\hline
		Dubrovnik \cite{li2010location} &  6,044 & 1.89M & 800 \\ 
		RobotCar Seasons \cite{sattler2018benchmarking}&  20,862 & 6.77M & 11,934 \\
		Aachen Day-Night \cite{sattler2018benchmarking} & 4,328 & 1.65M & 922 \\
		SF-0 \cite{li2012worldwide,sattler2017large} &  610,773 & 30M & 442 \\ \Xhline{1pt}
	\end{tabular}	
	\vspace{-2em}
\end{table} 
\subsection{Implementation Details}
For the Dubrovnik dataset, we use the same 10k general visual vocabulary trained by \cite{sattler2012improving}. For the RobotCar Seasons and Aachen Day-Night datasets, we train a specific 10k visual vocabulary on all upright RootSIFT descriptors found in 1000 randomly selected database images in the reference SfM model. For the large-scale SF-0 dataset, we train a 50k specific visual vocabulary on all integer mean RootSIFT descriptors. For the Dubrovnik and RobotCar Seasons datasets, we set $B = 64$, $\tau = 19$ and $\alpha = 0.8$ for feature-wise filtering. In the visibility-wise filtering step, we set $k=20$ and $k_1 = 100$. In the geometry-wise filtering step, we set $N = 100$. For the Aachen Day-Night dataset, we find that by keeping other parameters unchanged and  setting $\tau = 16$ and $k_1 = 50$ can obtain sufficient correct 2D-3D matches. Due to dramatically different characteristics between large-scale SF-0 and the above three medium-scale datasets, we adjust $B$ to 128, $\tau$ to 32 and $\alpha$ to 0.4 accordingly. For computing the auxiliary camera pose and the final camera pose, we run both 1000 RANSAC iterations. For a fair comparison on the Dubrovnik dataset, we use a threshold of 4 pixels for final pose estimation. For a fair comparison on the RobotCar Seasons and Aachen Day-Night dataset, we use a 3-point pose solver to compute the auxiliary camera pose and a threshold of 4 pixels for final pose estimation. All experiments were conducted with a single CPU thread on a PC with an Intel i7-6800K CPU with 3.40 GHz and 32 GB RAM.

\subsection{Comparison with State-of-the-art}
\begin{table}[!t]
	\centering
	\small
	\caption{The comparison between our method and state-of-the-art methods on the Dubrovnik dataset.} \label{table::compare_dubro}
	\begin{tabular}{lcccc}
		\Xhline{1pt}
		\multirow{2}{*}{Method} & \multicolumn{3}{c}{Error Quartiles {[}m{]}} & \multirow{2}{*}{ \begin{tabular}[c]{@{}c@{}}Localized \\ images\end{tabular}} \\ \cline{2-4}
		& $25\%$ & $50\%$ & $75\%$ &   \\ \hline 
		EGM	& 0.24 & 0.70 &2.67 & 794   \\ 
		TC	& \textbf{0.22} & 1.07 &2.99 & \textbf{800} \\ 
		AS & 0.40 & 1.40 &5.30 & 796 \\ 
		Our method &\textbf{0.22} & \textbf{0.64} & \textbf{2.16} & 794\\ \Xhline{1pt}
	\end{tabular}
	\vspace{-1em}
\end{table}
\begin{table}[!t]
	\centering
	\small
	\caption{The percentage of query images localized within three pose accuracy intervals of our proposed method compared with state-of-the-art localization methods on the RobotCar Seasons and Aachen Day-Night datasets. \textcolor{red}{\textbf{red}} and \textcolor{blue}{\textbf{blue}} represent the \textcolor{red}{\textbf{best}} and \textcolor{blue}{\textbf{second-best}} methods, and the asterisk symbol represents using knowledge about the gravity direction.}  \label{table::compare_robot}
	\begin{tabular}{l|c|c}
		\Xhline{1pt}
		\multicolumn{3}{c}{\textbf{RobotCar Seasons}} \\ \hline
		& All Day & All Night \\ \cline{2-3}
		\multicolumn{1}{r|}{m} & .25 / 0.5 / 5.0 & .25 / 0.5 / 5.0 \\ 
		\multicolumn{1}{r|}{deg} & 2 / 5 / 10 & 2 / 5 / 10 \\ \hline
		AS & 35.6 / 67.9 / 90.4 & 0.9 / 2.1 / 4.3 \\ 
		DenseVLAD & 7.7 / 31.3 / 91.2 & 1.0 / 4.5 / \textcolor{blue}{\textbf{22.7}} \\ 
		NetVLAD  & 6.4 / 26.3 / 91.0  & 0.4 / 2.3 / 16.0 \\ 
		$\text{CSL}^{*}$  & 45.3 / 73.5 / 90.1 & 0.6 / 2.6 / 7.2 \\ 
		$\text{SMC}^{*}$  & \textcolor{red}{\textbf{50.6}} / \textcolor{red}{\textbf{79.8}} / \textcolor{red}{\textbf{95.1}} & \textcolor{red}{\textbf{7.6}} / \textcolor{red}{\textbf{21.5}} / \textcolor{red}{\textbf{45.4}} \\ 
		Our method & \textcolor{blue}{\textbf{48.0}} / \textcolor{blue}{\textbf{78.0}} / \textcolor{blue}{\textbf{94.2}}  & \textcolor{blue}{\textbf{3.4}} / \textcolor{blue}{\textbf{9.5}} / 17.0\\ \Xhline{1pt}
		\multicolumn{3}{c}{\textbf{Aachen Day-Night}}\\\hline 
		& Day & Night \\ \cline{2-3}
		\multicolumn{1}{r|}{m} & .25 / 0.5 / 5.0 & .25 / 0.5 / 5.0 \\ 
		\multicolumn{1}{r|}{deg} & 2 / 5 / 10 & 2 / 5 / 10 \\ \hline
		AS & \textcolor{blue}{\textbf{53.7}} / \textcolor{blue}{\textbf{83.7}} / \textcolor{red}{\textbf{96.6}} & 19.4 / 30.6 / 43.9 \\ 
		DenseVLAD & 0.0/ 0.1 / 22.8 & 0.0/ 2.0 / 14.3 \\ 
		NetVLAD  & 0.0 / 0.2 / 18.9  & 0.0 / 2.0 / 12.2 \\ 
		$\text{CSL}^{*}$  & 52.3 / 80.0 / 94.3 & \textcolor{blue}{\textbf{24.5}} / \textcolor{blue}{\textbf{33.7}} / \textcolor{blue}{\textbf{49.0}} \\ 
		$\text{SMC}^{*}$ & - & - \\ 
		Our method & \textcolor{red}{\textbf{76.7}} / \textcolor{red}{\textbf{88.6}} / \textcolor{blue}{\textbf{95.8}}  & \textcolor{red}{\textbf{25.5}} / \textcolor{red}{\textbf{38.8}} / \textcolor{red}{\textbf{54.1}}\\ \Xhline{1pt}
	\end{tabular}
	\vspace{-2em}
\end{table}
On the Dubrovnik dataset, we compare against three prior-free state-of-the-art approaches: Efficient Global Matching (EGM) \cite{liu2017efficient}, Active Search (AS) \cite{sattler2012improving} and Toroidal Constraint (TC) \cite{camposeco2017toroidal}. On the other three datasets in which images are captured on the street, we include the comparison with approaches that use the knowledge about gravity direction. Concretely, we compare with City-scale Localization (CSL) \cite{svarm2017city}, Camera Pose Voting (CPV) \cite{zeisl2015camera} and Semantic Match Consistency (SMC) \cite{toft2018semantic}. For comprehensiveness, we also compare with two retrieval-based approaches, namely DenseVLAD \cite{arandjelovic2013all} and NetVLAD \cite{arandjelovic2016netvlad}. \\\vspace
{-0.6em}

\noindent \textbf{Evaluation on medium-scale datasets.} Table \ref{table::compare_dubro} shows the comparison on the Dubrovnik dataset. As can be seen, our method outperforms state-of-the-art methods in localization accuracy. In the meantime, we maintain a very competitive effectiveness, i.e., the number of successfully localized query images. Table \ref{table::compare_robot} shows the percentage of query images localized within three pose accuracy intervals of our proposed method compared with state-of-the-art localization methods on the RobotCar Seasons and Aachen Day-Night datasets. Our method achieves the second best localization performance on the RobotCar Seasons dataset. Interestingly, our method significantly outperforms CSL that requires prior knowledge about the gravity direction. SMC relies on a neural network for semantic segmentation. Note that the training data used in SMC includes several manually labelled images from the original RobotCar dataset \cite{maddern20171}. On the Aachen Day-Night dataset, our method achieves the best localization performance in most cases. \\\vspace{-0.8em}

\noindent \textbf{Memory consumption.} We also investigate the memory consumption required by our method and other methods. Without losing generality, we only compare against AS which is the most memory-efficient in state-of-the-art structure-based localization methods. Table \ref{table::storage} shows the detailed comparison. Comparing with AS, our method requires significantly lower memory consumption. The reason for the memory reduction is that our method only needs to store a compact binary signature (8-bytes when $B=64$) per visual word for each 3D point. While AS needs to store an integer mean (128-bytes) of SIFT descriptors per visual word for each 3D point. Overall speaking, our method is memory-efficient and achieves very competitive localization performance on medium-scale datasets. \\\vspace{-0.6em}

\begin{figure}[!t]
	\centering
	\includegraphics[width=0.4\textwidth]{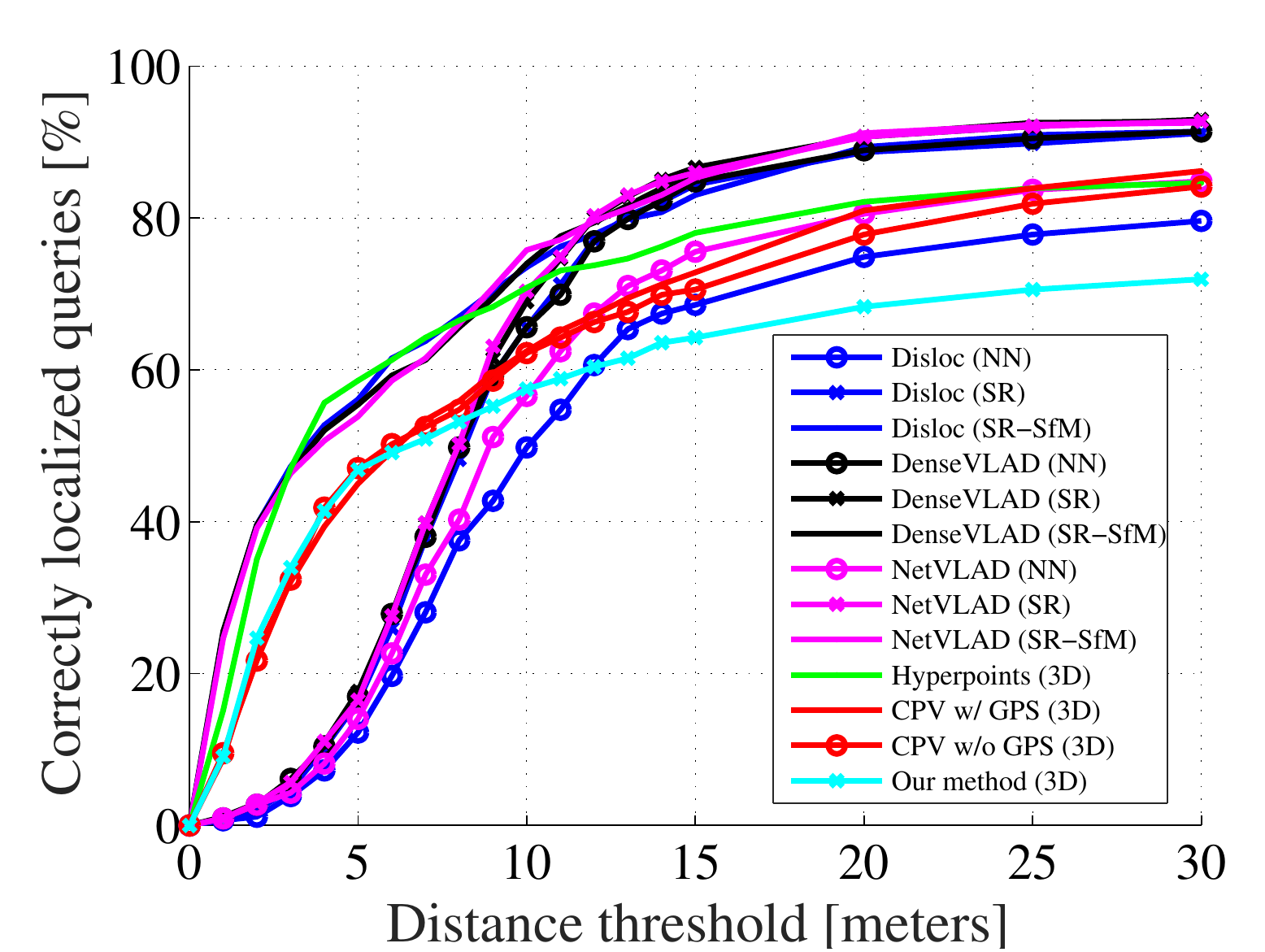}
	\caption{The experimental results on the SF-0 dataset.}\label{fig::sf}
	\vspace{-1.5em}
\end{figure}

\noindent \textbf{Evaluation on large-scale SF-0.} Fig. \ref{fig::sf} shows the results on the SF-0 dataset. We mainly compare with two structure-based methods: CPV and Hyperpoints \cite{sattler2015hyperpoints}. Note that the SR-SfM scheme in Fig. \ref{fig::sf} usually takes several minutes to process one query image. Comparing with CPV using full descriptors, our method achieves competitive results for thresholds of 5m or less. Yet, our method does not perform better than Hyperpoints, in which a fine vocabulary is used and more suitable for large-scale location recognition problems. In addition, using the GPS tags available in SF-0 would be beneficial to remedy the drawback of our method for coarse-level localization (5$\sim$30m).
\begin{table}[!t]
	\centering
	\small
	\caption{The memory consumption (in GB) comparison between our method and other state-of-the-art methods.} \label{table::storage}
	\vspace{-0.5em}
	\begin{tabular}{l|c|c|c}
		\Xhline{1pt}
		\multirow{3}{*}{Method} & \multicolumn{3}{c}{Memory Consumption} \\ \cline{2-4} 
		& Dubrovnik & RobotCar & Aachen\\ \hline 
		AS & 0.75 & 2.72 & 0.76 \\ 
		Our method & \textbf{0.14} & \textbf{0.52}  & \textbf{0.14} \\ \Xhline{1pt}
	\end{tabular}
	\vspace{-1.2em}
\end{table} 

\subsection{Ablation Study}
We conduct an ablation study on the Dubrovnik dataset to evaluate the impact of key components in our method. The match score threshold in the two-step match selection method is heavily related to the the bilateral Hamming ratio test. For simplicity, we arguably evaluate these two components together. To this end, we first implement a baseline voting method that filters wrong matches established from binary signatures. In the baseline implementation, a match is evaluated by Eq. \ref{eq::final_score}. Then, we select all matches from top-20 ranked database images for computing the auxiliary camera pose, and we select all matches from top-100 ranked database images to obtain the visibility-wise match pool. Other components in our method remain unchanged. We test with multiple Hamming distance thresholds in Eq. \ref{eq::final_score}, and the baseline implementation achieves the best performance when setting the threshold to 11. As shown in Table \ref{table::ablation_voting}, our method can localize 16 more query images than the baseline implementation. 
This indicates that the combination of the bilateral Hamming ratio test and the two-step match selection method is beneficial for better filtering.

We also conduct an experiment to investigate the impact of the quality-aware spatial reconfiguration (QSR) method and the principal focal length estimation (PFL) in Section \ref{Section:geometry}. We first disable QSR and select the same number of \emph{VFC} and \emph{VFC-I} matches as when QSR enabled. Note that the matches in QSR disabled are selected with the largest match scores. As shown in Table \ref{table::ablation_voting}, QSR significantly improves the localization accuracy. This indicates that obtaining a set of uniformly distributed matches before RANSAC-based pose estimation is essential for accurate IBL. To examine the benefit of PFL, we conduct an experiment with traditional RANSAC scheme when computing the auxiliary camera pose, \ie the best camera pose is the one with largest number of inliers. We can see that PFL also significantly improves the localization accuracy. This indicates that the auxiliary camera pose selected with PFL is more robust to apply geometry-wise match filtering.
\begin{table}[!t]
	\centering
		\small
	\caption{The ablation study conducted on the Dubrovnik dataset.  } \label{table::ablation_voting}
	\vspace{-0.5em}
	\begin{tabular}{lccccc}
		\Xhline{1pt}
		\multirow{2}{*}{Setting} & \multicolumn{3}{c}{Error quartiles {[}m{]}} & \multirow{2}{*}{\begin{tabular}[c]{@{}c@{}}Localized \\ images\end{tabular}} \\ \cline{2-4}
		& 25\% & 50\% & 75\% &  \\ \hline
		Baseline Voting& 0.25 & 0.69& 2.19  & 778 \\ 
		w/o QSR  & 0.26 & 0.74& 2.53  & 793 \\ 
		w/o  PFL & 0.31 & 0.80& 2.70  & \textbf{794}  \\
		Our full method&\textbf{0.22}  &\textbf{0.64} &\textbf{2.16}  & \textbf{794} \\ \Xhline{1pt}
	\end{tabular}%
	\vspace{-1.5em}
\end{table}
\section{Conclusion}\label{conclusion}
In this paper, we have presented a cascaded parallel filtering method for memory-efficient image-based localization. Our method contains a cascade of feature-, visibility- and geometry-based filters, in which two parallel criteria are applied for preserving correct matches and obtaining high quality matches. The localization accuracy is improved by quality-aware spatial reconfiguration and principal focal length methods. Comprehensive experiments on challenging real-world datasets demonstrate the benefit of our method. Further improvements could be achieved by incorporating CNN-based feature descriptors \cite{Dusmanu2019D2} or hierarchical localization schemes \cite{Sarlin2018From}. \\ \vspace{-0.8em}
	
	\noindent \textbf{Acknowledgements: } This research was supported by Singapore Ministry of Education Tier-2 Fund MOE2016-T2-2-057(S) and the National Research Foundation, Prime Minister's Office, Singapore under its International Research Centres in Singapore Funding Initiative.

\end{document}